\documentclass[sigconf]{acmart}
\AtBeginDocument{%
  }

\setcopyright{acmlicensed}
\copyrightyear{2018}
\acmYear{2018}
\acmDOI{XXXXXXX.XXXXXXX}
\acmConference[ICMR' 25]{Make sure to enter the correct conference title from your rights confirmation email}{June 30--July 03, 2025}{Chicago, USA}
\acmISBN{978-1-4503-XXXX-X/2018/06}

\acmSubmissionID{503}
\usepackage{colortbl} 
\usepackage{tabularx}

\begin{document}


\title{SSD-Poser: Avatar Pose Estimation with State Space Duality from Sparse Observations}

\author{Shuting Zhao}
\email{zhaoshuting@fudan.edu.cn}
\orcid{0009-0006-0996-0835}
\affiliation{%
  \institution{Academy for Engineering \& Technology, Fudan University}
  \city{Shanghai}
  \country{China}
}

\author{Linxin Bai}
\email{24210860025@m.fudan.edu.cn}
\affiliation{%
  \institution{Academy for Engineering \& Technology, Fudan University}
  \city{Shanghai}
  \country{China}
}

\author{Liangjing Shao}
\email{ljshao22@m.fudan.edu.cn}
\affiliation{%
  \institution{Academy for Engineering \& Technology, Fudan University}
  \city{Shanghai}
  \country{China}
}

\author{Ye Zhang}
\authornote{Corresponding author.}
\email{yezhang@sspu.edu.cn}
\affiliation{
  \institution{College of Vocational and Technical Teacher Education, Shanghai Polytechnic University}
  \city{Shanghai}
  \country{China}
}

\author{Xinrong Chen}
\email{chenxinrong@fudan.edu.cn}
\authornotemark[1]
\affiliation{
  \institution{Academy for Engineering \& Technology, Fudan University}
  \city{Shanghai}
  \country{China}
}



\renewcommand{\shortauthors}{Shuting Zhao et al.}
\settopmatter{printacmref=false} 
\renewcommand\footnotetextcopyrightpermission[1]{}

\begin{abstract}
The growing applications of AR/VR increase the demand for real-time full-body pose estimation from Head-Mounted Displays (HMDs). Although HMDs provide joint signals from the head and hands, reconstructing a full-body pose remains challenging due to the unconstrained lower body. Recent advancements often rely on conventional neural networks and generative models to improve performance in this task, such as Transformers and diffusion models. However, these approaches struggle to strike a balance between achieving precise pose reconstruction and maintaining fast inference speed. To overcome these challenges, a lightweight and efficient model, SSD-Poser, is designed for robust full-body motion estimation from sparse observations. SSD-Poser incorporates a well-designed hybrid encoder, State Space Attention Encoders, to adapt the state space duality to complex motion poses and enable real-time realistic pose reconstruction. Moreover, a Frequency-Aware Decoder is introduced to mitigate jitter caused by variable-frequency motion signals, remarkably enhancing the motion smoothness. Comprehensive experiments on the AMASS dataset demonstrate that SSD-Poser achieves exceptional accuracy and computational efficiency, showing outstanding inference efficiency compared to state-of-the-art methods.
\end{abstract}





\maketitle

\section{Introduction}
The estimation of full-body pose in AR/VR scenes garners significant attention due to its wide applications, including humanoid robot control~\cite{1} and virtual character reconstruction. In these scenarios, Head-Mounted Devices (HMDs) are primarily used to capture the input signals, capturing sparse sensory data from the user's head and hands. However, there are significant challenges in full-body motion tracking based on sparse signal inputs, especially in capturing unconstrained lower-body movements.

Previous works~\cite{6,3,5,18,7,2} on full-body pose estimation from sparse sensors primarily rely on conventional neural networks or generative models. While these traditional neural networks, especially Transformers, excel at modeling context relationships and extracting long-range dependencies, they often face limitations in quadratic computational complexity and dynamic feature modeling. Conversely, generative models that produce smoother motion sequences and finer pose details, including diffusion models, have high computational demands and slow inference speeds, which limit the ability to reconstruct full-body motion in real time.


\begin{figure}[t]
  \centering
  \includegraphics[width=1.0\linewidth]{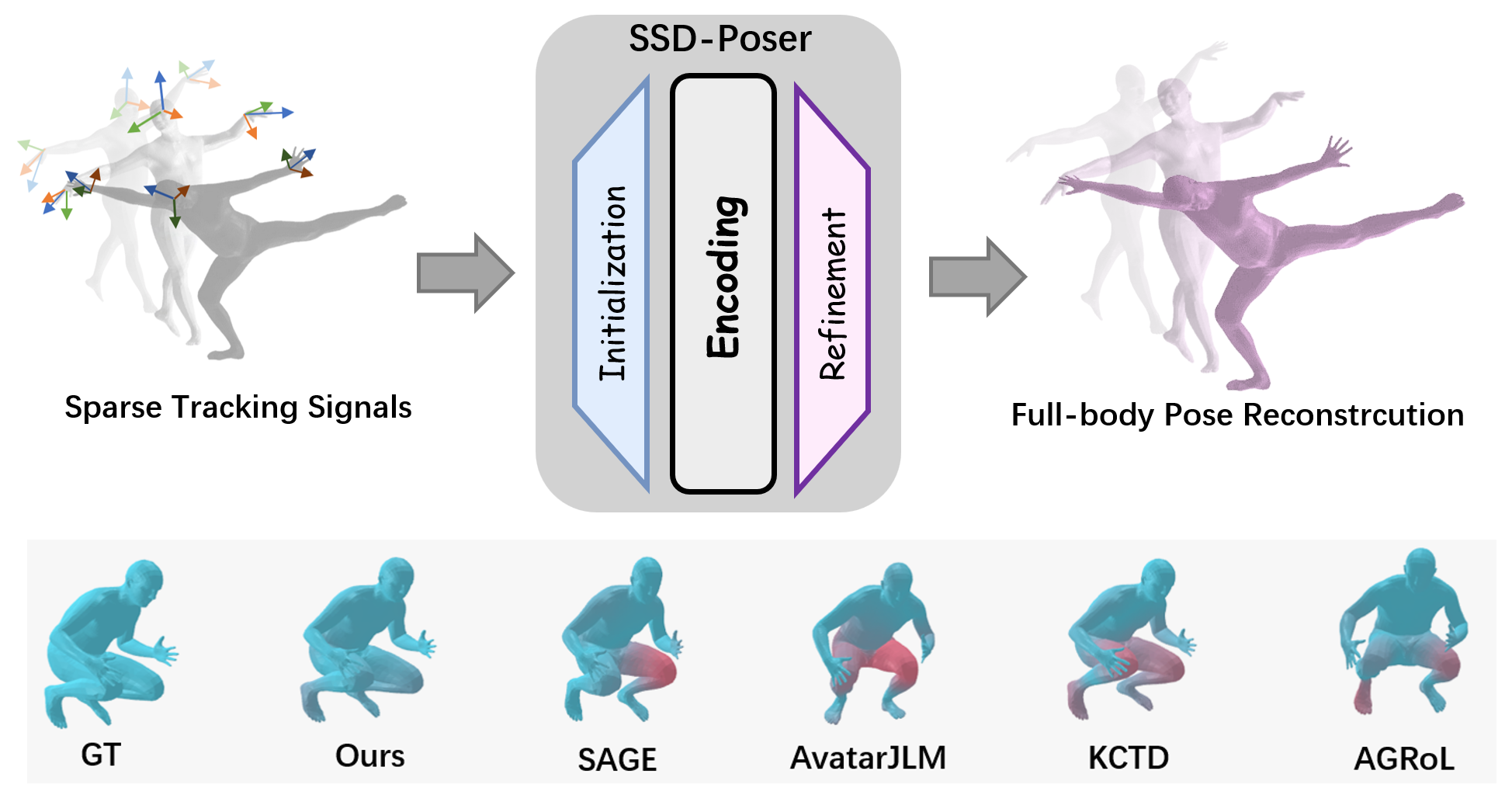}
  \caption{Our method takes sparse observations as inputs and reconstructs full-body motion as output, outperforming the state-of-the-art methods.}
\label{fig0}
\end{figure}

\begin{figure}[t]
\centering
\includegraphics[width=3.5in]{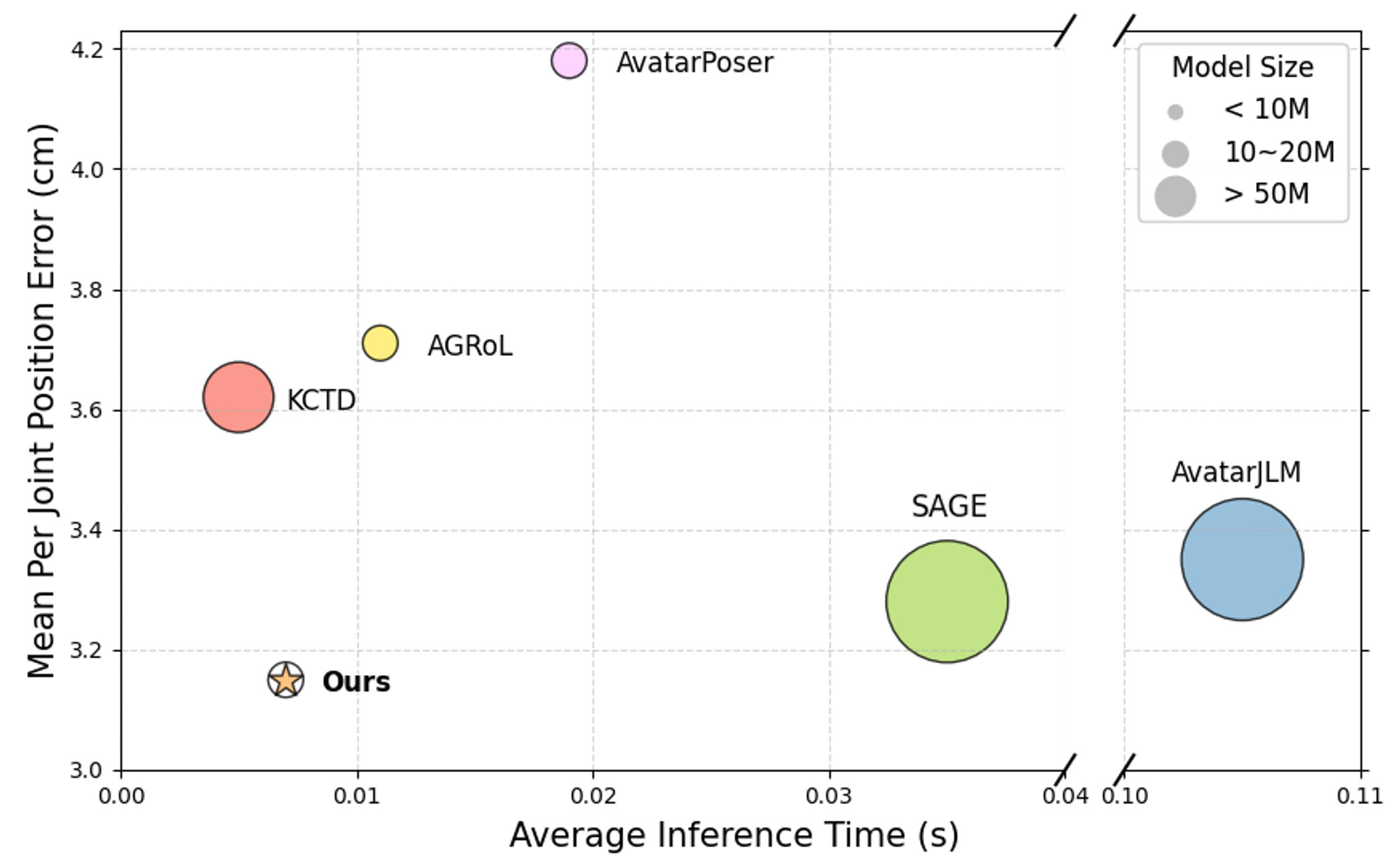}%
\hfil
\caption{Comparison of our approach with state-of-the-art methods in terms of overall performance. 
Our method achieves the smallest average position error, the fast inference speed, and maintains a lightweight model architecture.}
\label{LC}
\end{figure}

Recent advancements in State Space Models (SSMs), particularly the State Space Duality (SSD) framework~\cite{8}, have gained prominence for processing long-term sequential data. The SSD framework is not only expert in dynamic system modeling but also enables efficient training on hardware through parallel scanning algorithms. This inspires us to employ the SSD framework to realize efficient and realistic full-body motion reconstruction.

However, when the SSD framework is applied to full-body pose estimation, some challenges need to be addressed. First, this framework strongly relies on input signals and is susceptible to sparse observations, which often leads to inaccurate inference of hidden states in the observation equations. Second, the framework struggles to capture complex contextual features, such as inter-joint correlations in periodic motion, especially when compared to Transformers~\cite{30}.

To address these challenges, a lightweight and highly efficient model based on the SSD framework, SSD-Poser, is proposed specifically for full-body pose estimation. SSD-Poser consists of three stages, namely feature initialization stage, spatiotemporal encoding stage, and pose refinement stage. In the first stage, a Base Feature Extractor (BFE) is employed, which uses a simple linear layer to initialize motion features from sparse tracking signals. In the second stage, a well-designed hybrid module, State Space Attention Encoder (SSAE), is leveraged. SSAE utilizes the powerful capabilities of the SSD framework to reduce computational overhead and extract dynamic spatiotemporal features, while simultaneously incorporating Transformers to capture intricate contextual information and long-range dependencies. Furthermore, in the third stage, a Frequency-Aware Decoder (FAD) is designed to process variable-frequency signals and refine human poses, which minimizes motion jitter and produces highly realistic full-body poses.

Experiments are conducted on the AMASS dataset~\cite{10} with large-scale motion capture, and the proposed SSD-Poser exhibits significant improvements in both inference speed and accuracy, particularly in lower-body motion prediction compared to previous state-of-the-art methods. In summary, the contributions can be summarized as follows:
\begin{itemize}
\item A novel model based on the SSD framework is proposed for full-body motion estimation using sparse signals from HMDs, which features a lightweight and efficient network architecture.

\item The State Space Attention Encoder (SSAE) is designed to efficiently capture both dynamic spatiotemporal features and long-distance contextual dependencies while alleviating computational cost. 

\item The Frequency-Aware Decoder (FAD) is introduced to address variable-frequency signals and refine human poses, enhancing motion smoothness and delivering highly realistic avatar poses. 

\item Comprehensive experiments and visualization results indicate that SSD-Poser achieves state-of-the-art performance, providing an innovative and efficient backbone for full-body pose estimation tasks. 
\end{itemize}


\section{Related Works}
\subsection{Human Pose Estimation from Sparse Observations }
The estimation of full-body poses from sparse observations attracted significant attention within the research community~\cite{51,52,53,54,4,17,12,15,13,3,5,2,18,17,7}.  The methods for acquiring sparse observations are primarily categorized into three types, 6 inertial measurement units (IMUs), 4 IMUs, or head-mounted devices (HMDs). Previous works~\cite{4,12,15} relied on 6 IMUs to track full-body motion. DIP~\cite{4} employed a deep learning model to explore the time dependencies and temporal pose priors, while PIP~\cite{12} incorporated physical constraints to improve task performance. Furthermore, LoBSTr~\cite{13} utilized 4 IMUs to capture sparse tracking signals, specifically from the head, two hands, and pelvis.

In the context of AR/VR, HMDs were widely adopted, but they only provided three tracking signals from the head and hands. AvatarPoser~\cite{3} and AvatarJLM~\cite{5} built Transformer-based architectures to predict full-body pose from these 3 sparse inputs. Moreover, KCTD~\cite{2} introduced a MLP-based model incorporating kinematic constraints for the same task. Additionally, some methods approached the problem as a synthesis task. For example, FLAG~\cite{18} and VAE-HMD~\cite{17} employed variational auto-encoders (VAEs) and flow-based techniques, respectively, while SAGE~\cite{7} utilized a diffusion model and a VQ-VAE~\cite{16} for avatar generation. 

Existing methods encounter challenges in precisely estimating full-body motion or achieving rapid inference speeds. To address this, a lightweight and efficient network is introduced, combining an SSD-based hybrid encoder with a frequency-aware decoder, which enhances both accuracy and efficiency in full-body pose prediction. 

\subsection{MLP and Transformer for pose prediction  }
The multi-layer perceptrons (MLPs) and Transformer architectures~\cite{19} were widely applied in full-body pose estimation due to their robust feature extraction capabilities. Recently, a long-term MLP block~\cite{2,6} was introduced as a backbone for effectively capturing temporal information. Additionally, Transformer-based networks~\cite{3,5,11} leveraged self-attention mechanisms to capture body joint correlations and long-range dependencies. As a result, both MLPs and Transformer blocks demonstrated significant potential for full-body motion tracking tasks.

\subsection{State Space Model }
State Space Models (SSMs)~\cite{20,22,25} recently gained significant attention for sequential data processing by connecting inputs and outputs through latent states that capture the underlying dynamics of a system~\cite{28}. One of the main advantages of SSMs was their ability to scale linearly with sequence length, making them highly effective for modeling long sequence, which attracted substantial research interest. For example, the structured state-space sequence model (S4)~\cite{20} was introduced, utilizing a diagonal structure to normalize parameters for time series analysis. The subsequent development of the S5 layer~\cite{25} and the gated state space layer (GSS)~\cite{46} further enhanced the efficiency of S4 by improving parallel scanning and incorporating gated mechanisms, respectively. More recently, Mamba~\cite{9} and Mamba-2~\cite{8} shown competitive performance, faster inference speeds, and linear scaling with constant memory usage. As a result, various Mamba-based approaches emerged within computer vision~\cite{23,24,26,27}. In this study, we explore the core framework of Mamba-2, known as State Space Duality (SSD), to track full-body motion from sparse observations, which can be established as a simple yet robust foundation for future developments through targeted design integration.

\begin{figure*}[!t]
\centering
\includegraphics[width=7.3in]{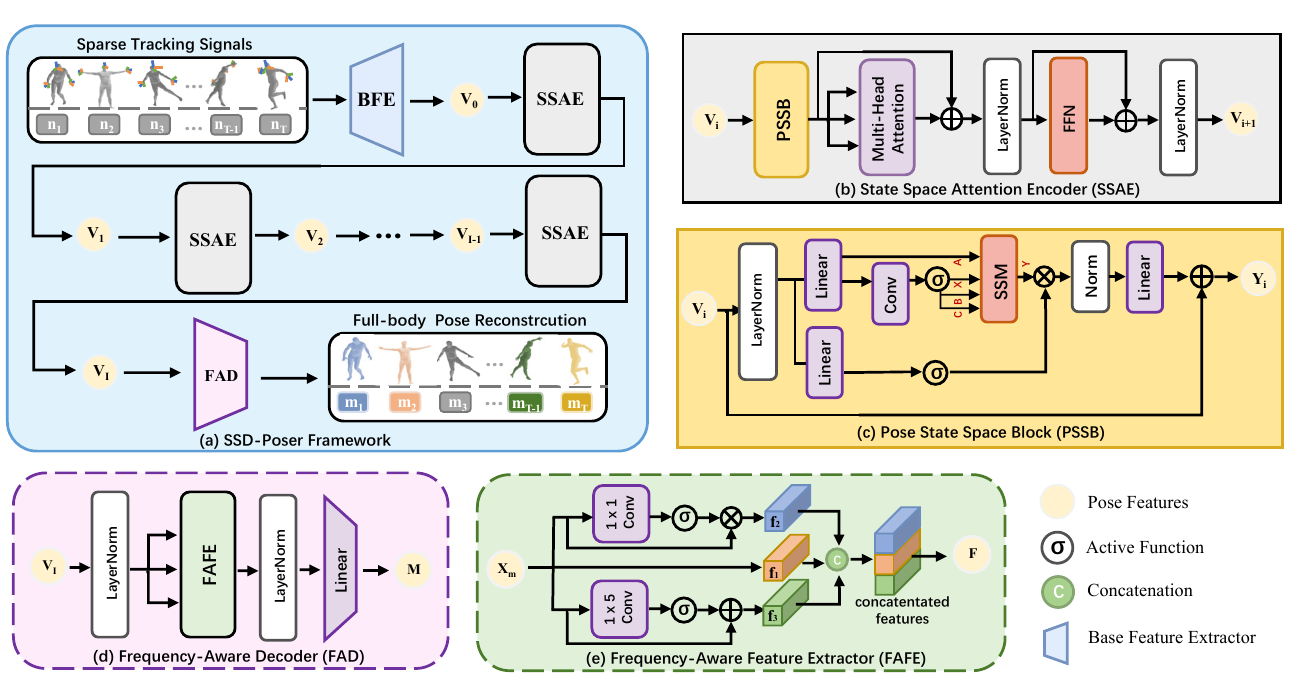}%
\hfil
\caption{The overall framework of this work. (a) The framework of the proposed SSD-Poser model. (b) The framework of the State Space Attention Encoder (SSAE). (c) The framework of the Pose State Space Block (PSSB). (d) The framework of the Frequency-Aware Decoder (FAD). (e) The framework of the Frequency-Aware Feature Extractor (FAFE).}
\label{fig1}
\end{figure*}

\section{Preliminaries}
SSMs represent a class of mathematical frameworks widely used to describe dynamic systems, with extensive applications in signal processing and time-series analysis. Recognizing the potential of SSMs for content-aware modeling and efficient computation, Dao et al.~\cite{8} proposed the SSD framework. This framework introduces a novel matrix-based computation structure, which integrates the linear recurrence properties of SSMs with a quadratic dual formulation. Specifically, the SSM formulation under the SSD framework is expressed as:
\begin{equation}
\begin{gathered}
\label{eq.5}
y_t=\sum_{i=0}^t C_t^T A_{t: i}^X B_i x_i, \\
y=S S M(A, B, C)(x)=H x.
\end{gathered}
\end{equation}
The Eq.(\ref{eq.5}) delineates the transformation of the input sequence $x\in \mathbb{R}^{T \times D}$ into
the output sequence $y\in \mathbb{R}^{T \times D}$, facilitated by the parameter matrices $A\in \mathbb{R}^{T \times N}$, $B \in \mathbb{R}^{T \times N}$, $C \in \mathbb{R}^{T \times N}$, and $H \in \mathbb{R}^{T \times T}$. In this context, $A_{t: i}^X $ represents the sequential product of $\mathrm{A}$ terms from $i$ + 1 to $t$, and $H$ is defined as: 
\begin{equation}
H_{j i}:=C_j^{\top} A_j \cdots A_{i+1} B_i.
\end{equation}
When $A_i $ is reduced to a scalar, Eq.\eqref{eq.5} can be reformulated as:
\begin{equation}
{y}={Hx}={F} \cdot\left({C}^{{T}} {~B}\right) {x},
\end{equation}
where 
\begin{equation}
{F}_{{ij}}= \begin{cases}A_{i+1} \times \cdots \times A_j & i>j \\ 1 & i=j \\ 0 & i<j\end{cases}.
\end{equation}


Moreover, the SSM operator SSM(A, B, C) can also be defined as the sequence transformation:
\begin{equation}
\begin{gathered}
h_t=A_t h_{t-1}+{B}_t x_t,\\
y_t={C}_t^T h_t,
\end{gathered}
\end{equation}
where $h_{t}$ is an implicit latent state, which bridges the input $x_{t}$ and the output $y_{t}$.

The SSD framework offers a distinct advantage for sequential data processing, supporting efficient parallel computation and enabling the modeling of dynamic features. In this work, the potential of the SSD framework is explored to reconstruct full-body pose with pose-specific designs, which is established as a lightweight yet effective baseline for future research.

\section{Method}

\subsection{Problem Formulation  }
This task aims to predict realistic full-body motion using sparse IMU signals, with HMDs serving as the main source of sparse tracking inputs.

Sparse tracking inputs are denoted as $\mathbf{N}=\left\{\mathbf{n}_i\right\}_{i=1}^T \in \mathbb{R}^{T \times C}$, where $T$ represents the number of time frames, and $C$ denotes the dimension of input signals. Each $\mathbf{n}_i$ consists of three components corresponding to the features of the head and both hands. Based on ~\cite{3}, each component includes a 3D position $\mathbf{p}^{1\times3}$, a 6D rotation $\theta^{1\times6}$, a linear velocity $\mathbf{v}^{1\times3}$, and an angular velocity $\omega^{1\times6}$. Consequently, each input $x_i$ can be expressed as:
\begin{equation}
\begin{gathered}
\mathbf{n}_i^{1 \times 18 \times 3}=\left[\left\{\mathbf{p}_i^1, \mathbf{v}_i^1,
\theta_i^1, \omega_i^1\right\}, \ldots,\left\{\mathbf{p}_i^3, \mathbf{v}_i^3, \theta_i^3, \omega_i^3\right\}\right].
\end{gathered}
\end{equation}
As a result, the dimension of input signals $C$ equals 54.

To represent the human pose, the SMPL~\cite{29} model is commonly employed, and the output full-body poses are represented as $\mathbf{M}=\left\{\mathbf{m}_i\right\}_{i=1}^T \in \mathbb{R}^{T \times S}$. In this task, only the first 22 joints of the SMPL model are used, while the pose of the fingers is disregarded. Therefore, the dimension of the final output is S = 22 × 6 = 132.

\subsection{Overall Framework  }
The overall framework of the proposed SSD-Poser model is illustrated in Fig \ref{fig1}. The model comprises three stages, namely feature initialization stage, spatiotemporal encoding stage, and pose refinement stage. Given the sparse tracking signals $\mathbf{N} \in \mathbb{R}^{T \times C}$, we first use the Base Feature Extractor (BFE) as a single linear layer to generate initial pose features ${V_0} \in \mathbb{R}^{T \times E}$. To further extract deep spatiotemporal information, State Space Attention Encoders (SSAEs) comprising several SSAE modules are employed in the second stage. After the pose features $V_0$ passes through the $i$-th SSAE, the  ${M_i} \in \mathbb{R}^{T \times E}$ is obtained, where ${i} \in \left\{1,2,...,I\right\}$. Here, $T$ denotes the number of frames, while $C$ and $E$ represent the feature dimensions of the sparse and initialized signals, respectively. Finally, Frequency-Aware Decoder (FAD) is introduced to refine full-body pose reconstruction, significantly improving the motion smoothness and realism.

\begin{table*}[h]
    \centering
    \caption{Comparison of our approach with state-of-the-art methods on three subsets of AMASS in the first setting. The best results are in bold, and the second-best results are underlined.}
    \label{tab:Table1}
    \begin{tabularx}{\textwidth}{l|>{\centering\arraybackslash}X>{\centering\arraybackslash}X>{\centering\arraybackslash}X>{\centering\arraybackslash}X>{\centering\arraybackslash}X>{\centering\arraybackslash}X>{\centering\arraybackslash}X>{\centering\arraybackslash}X}
    \toprule
     Method & MPJRE & MPJPE & MPJVE & Hand PE & Upper PE & Lower PE & Root PE & Jitter \\
     \midrule
     AvatarPoser~\cite{3} & 3.08 & 4.18 & 27.70 & 2.12 & 1.81 & 7.59 & 3.34 & 14.49 \\ 
     DAP~\cite{50} & 2.88 & 4.40 & 34.86 & 2.66 & 2.01 & 7.85 & 3.71 & 22.25 \\
     AGRoL~\cite{6} & 2.66 & 3.71 & \textbf{18.59} & 1.31 & 1.55 & 6.84 & 3.36 & \underline{7.26} \\
     KCTD~\cite{2} & 2.60 & 3.62 & 20.57 & 1.78 & 1.62 & 6.52 & 3.24 & 10.73 \\
     AvatarJLM~\cite{5} & 2.90 & 3.35 & 20.79 & 1.24 & 1.72 & 6.20 & 2.94 & 8.39 \\
     SAGE~\cite{7} & \underline{2.53} & \underline{3.28} & 20.62 & \underline{1.18} & \underline{1.39} & \underline{6.01} & \underline{2.95} & \textbf{6.55} \\
     \midrule
     \rowcolor[gray]{0.94} 
     \textbf{SSD-Poser (Ours)} & \textbf{2.41} & \textbf{3.15} & \underline{19.32} & \textbf{1.17} & \textbf{1.34} & \textbf{5.78} & \textbf{2.82} & 8.19 \\
     GT & 0 & 0 & 0 & 0 & 0 & 0 & 0 & 4.00 \\
    \bottomrule
    \end{tabularx}
\end{table*}

\begin{figure*}
    \centering
    \includegraphics[width=0.9\linewidth]{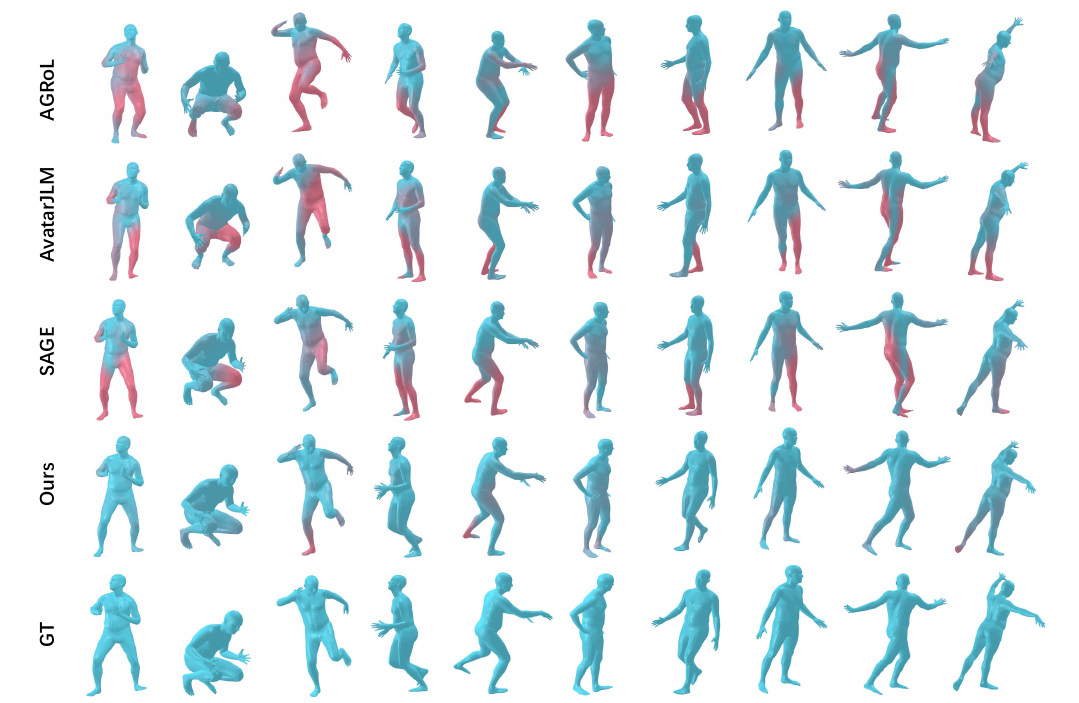}
    \caption{Visualization results of different actions compared with other state-of-the-art methods. Red regions indicate reconstruction errors, where wider or deeper red areas represent larger difference.}
    \label{fig2}
\end{figure*}


\subsection{State Space Attention Encoder}

To achieve realistic full-body pose estimation from sparse signals, numerous previous studies employ MLP-based or Transformer-based networks~\cite{2,5,6,11} to extract spatiotemporal features in poses. On the one hand, MLP-based networks for full-body pose estimation frequently encounter limited scalability and generalization. On the other hand, transformer-based networks, while effective for long-sequence modeling and context-rich feature extraction, demand substantial memory and quadratic computational complexity, which can hinder real-time performance required for full-body reconstruction. 

Recently, a novel effective framework named SSD~\cite{8} is proposed, which attracts attention due to its local feature extraction capabilities and computational efficiency. However, in terms of handling long-distance relationships, SSD still lags behind the performance of Transformer~\cite{30}.

To address these issues, we propose the State Space Attention Encoder (SSAE) for pose estimation. The hybrid framework SSAE combines the computational efficiency of SSD~\cite{8} with the long-range contextual modeling capability of Transformers~\cite{30}. As shown in Fig \ref{fig1} (b), SSAE consists of a Pose State Space Block (PSSB) and an Attention Module (AM)  incorporating a Muti-Head Attention block, a Feed-Forward Network (FFN)~\cite{19} and layer normalization (LN).

Given the input features $\mathrm{V_i}$, the PSSB is first applied to capture dynamic spatiotemporal information $\mathrm{Y}_{i}$. To address the lack of long-distance contextual dependencies, the AM is subsequently employed to produce the output $\mathrm{V_{i+1}}$:
\begin{equation}
\begin{gathered}
\mathrm{Y}_{i} = \mathrm{PSSB}\left(\mathrm{V_i}\right),\\
\mathrm{V}_{i+1} =\mathrm{AM}\left(\mathrm{Y}_{i}\right) ,\\
\end{gathered}
\end{equation}
In the structure of PSSB, illustrated in Fig \ref{fig1} (c), the input feature $V_i$ is processed through multiple branches, with the main branch modeled as a transformation from A, X, B, C to Y. Here, A, X, B and C are produced by the beginning projection comprising layer normalization, a linear layer, a convolution layer and a SiLU activation function. The complete process is detailed as follows:
\begin{equation}
\begin{gathered}
\mathrm{Y}_i^1 = \mathrm{SiLU}\left(\mathrm{Conv}\left(\mathrm{Linear}\left(\mathrm{LN}\left(\mathrm{V_i}\right)\right)\right)\right),\\
\mathrm{Y}_i^2 = \mathrm{SSM}\left(\mathrm{Y}_i^1+\mathrm{Linear}\left(\mathrm{LN}\left(\mathrm{V_i}\right)\right)\right),\\
\mathrm{Y}_i^3 = \mathrm{SiLU}\left(\mathrm{Linear}\left(\mathrm{LN}\left(\mathrm{V_i}\right)\right)\right),\\
\mathrm{Y}_{i} = \mathrm{Linear}\left(\mathrm{LN}\left(\mathrm{Y}_i^2\odot\mathrm{Y}_i^3\right)\right)+\mathrm{V_i}.
\end{gathered}
\end{equation}

 \subsection{Frequency-Aware Decoder}
Smoothness, as one of the key performance indicators in human motion tracking, has a significant impact on the realism and flexibility of avatar reconstruction. Most studies~\cite{3,50,5,7} focus on the precision of each joint, which may overlook the motion continuity induced by different motion frequencies. 

To achieve smooth and high-fidelity full-body motion reconstruction, the Frequency-Aware Decoder (FAD) is proposed to process temporal signals with varying frequencies. By separating the extraction of low and high-frequency features, FAD enables a more targeted learning process. Specifically, The low-frequency features focus on the broader, smoother aspects of human motion, while the high-frequency features capture the finer, rapid movements that add detail to the reconstruction. 

As illustrated in Fig \ref{fig1}(d), FAD comprises layer normalization, Frequency-Aware Feature Extractor (FAFE) and a linear layer, which can be denoted as follows:
\begin{equation}
\begin{gathered}
\mathrm{X}_{m} =\mathrm{LN}\left(\mathrm{V_I}\right),\\
\mathbf{M} =\mathrm{Linear}\left(\mathrm{LN}\left(\mathrm{FAFE}\left( \mathrm{X}_{m}\right)\right)\right).
\end{gathered}
\end{equation}
The core component of FAD is the FAFE, shown in Fig \ref{fig1}(e), which primarily adopts a residual structure combining a convolution layer and a $\mathrm{SiLU}$~\cite{32} activation function. Specifically, given the feature $\mathrm{X_m} $, the FAFE extracts low-frequency features through a convolution layer with a kernel size of $1\times1$ ($\mathrm{Conv_{1 \times 1}}$), while capturing high-frequency information using a convolution layer with a kernel size of $1\times5$ ($\mathrm{Conv_{1 \times 5}}$), which extends the temporal receptive field. This process is represented as follows: 
\begin{equation}
\begin{gathered}
{f_1}=\mathrm{X_m},\\
{f_2}=\mathrm{SiLU}\left(\mathrm{Conv_{1\times 1}}\left(\mathrm{X_L}\right)\right)\odot \mathrm{X_L},\\
{f_3}=\mathrm{SiLU}\left(\mathrm{Conv_{1\times 5}}\left(\mathrm{X_L}\right)\right)+\mathrm{X_L},\\
{F}={f_1} \oplus {f_2} \oplus {f_3},
\end{gathered}
\end{equation}
where $\odot$ denotes the Hadamard product, and $\oplus$ represents the concatenation operator. 

 \subsection{Loss Function }
The loss functions are adopted, which are local rotational loss ($L_\text{rot}$), positional loss ($L_\text{pos}$), and root joint orientation loss ($L_\text{ori}$) in line with prior work~\cite{3}. Rather than relying on the commonly used $L_1$ loss, the $L_2$ norm loss is used to impose stronger constraints, formulated  as follows:
\begin{equation}
\begin{gathered}
\text {L}_{rot}=\frac{1}{N} \sum_{i=1}^N \|{ R }_i-\hat{{R }_i} \|_2,\\
\text {L}_{pos}=\frac{1}{N} \sum_{i=1}^N \|{ P }_i-\hat{{P }_i} \|_2,\\
\text {L}_{ori}=\frac{1}{N} \sum_{i=1}^N \|{ O }_i-\hat{{O}_i} \|_2,
\end{gathered}
\end{equation}
where $\hat{}$ represents ground truth, and $R$, $P$, and $O$ denote the predicted joint features of local rotation, position, and orientation, respectively.

\begin{table*}[h]
\centering
\caption{Comparison of our approach with state-of-the-art methods on fourteen subsets of AMASS in the second setting.}
\label{tab:Table3}
\begin{tabularx}{\textwidth}{l|>{\centering\arraybackslash}X>{\centering\arraybackslash}X>{\centering\arraybackslash}X>{\centering\arraybackslash}X>{\centering\arraybackslash}X>{\centering\arraybackslash}X>{\centering\arraybackslash}X>{\centering\arraybackslash}X}
\toprule
Method & MPJRE & MPJPE & MPJVE & Hand PE & Upper PE & Lower PE & Root PE & Jitter \\
\midrule
AvatarJLM~\cite{5} & 3.14 & 3.39 & 15.75 & \textbf{0.69} & 1.48 & 6.13 & 3.04 & 5.33 \\
AGRoL~\cite{6} & 2.83 & 3.80 & 17.76 & 1.62 & 1.66 & 6.90 & 3.53 & 10.08 \\
AvatarPoser~\cite{3} & 2.72 & 3.37 & 21.00 & 2.12 & 1.63 & 5.87 & 2.90 & 10.24 \\
SAGE~\cite{7} & 2.41 & 2.95 & 16.94 & 1.15 & 1.28 & 5.37 & 2.74 & \textbf{5.27} \\
\midrule
\rowcolor[gray]{0.94}
\textbf{SSD-Poser (Ours)} & \textbf{2.18} & \textbf{2.67} & \textbf{15.25} & 1.07 & \textbf{1.19} & \textbf{4.80} & \textbf{2.45} & 6.73 \\
GT & 0 & 0 & 0 & 0 & 0 & 0 & 0 & 2.34 \\
\bottomrule
\end{tabularx}
\end{table*}

In summary, the overall loss function is expressed as:
\begin{equation}
\text{L}=\alpha \text{L}_ {rot} +\beta \text{L}_{pos}+\gamma \text{L}_{ori}.
\end{equation}
Here, the weights $\alpha$, $\beta$ and $\gamma$ are set to 1, 1 and 0.02, respectively.

\begin{table}[h]
\centering
\caption{Comparison of our approach with state-of-the-art methods in terms of parameter count and inference time.}
\label{tab:table2}
\begin{tabular}{l|cc}
\toprule  Methods & Params (M)  & Average Inference Time (s) \\ 
\midrule  AvatarPoser~\cite{3} & 4.125 & 0.019 \\
        DAP~\cite{50} & 15.56 & 0.068 \\
        AGRoL~\cite{6} & 7.48 & 0.011 \\
        KCTD~\cite{2} & 12.34 & 0.005 \\
        Avatar]LM~\cite{5} & 63.81 & 0.105 \\
        SAGE~\cite{7} & 137.35 & 0.035 \\
        \midrule
        \rowcolor[gray]{0.94} 
        \textbf{SSD-Poser (Ours)}& 7.34& 0.007 \\
\bottomrule
\end{tabular}
\end{table}

\section{Experiments}

\subsection{Dataset and Implementation Details}

The proposed method is trained and tested on the AMASS~\cite{10} dataset, a large human motion database, which unifies different motion capture datasets and is represented by SMPL~\cite{29} model parameters.

In the experiment, we design the SSAEs network with 4 blocks, each having a latent feature dimension of 256. To enhance performance, the input sequence length $T$ is set to 96 frames. The network is trained on an NVIDIA 4090 device with a batch size of 256. For parameter optimization, we employ the Adam solver~\cite{37} with a weight decay of 1e-5 during the training process. The learning rate is initialized to 3e-4 and decays to 3e-5 after 200000 iterations.

\begin{figure*}[h]
    \centering
    \includegraphics[width=0.9\linewidth]{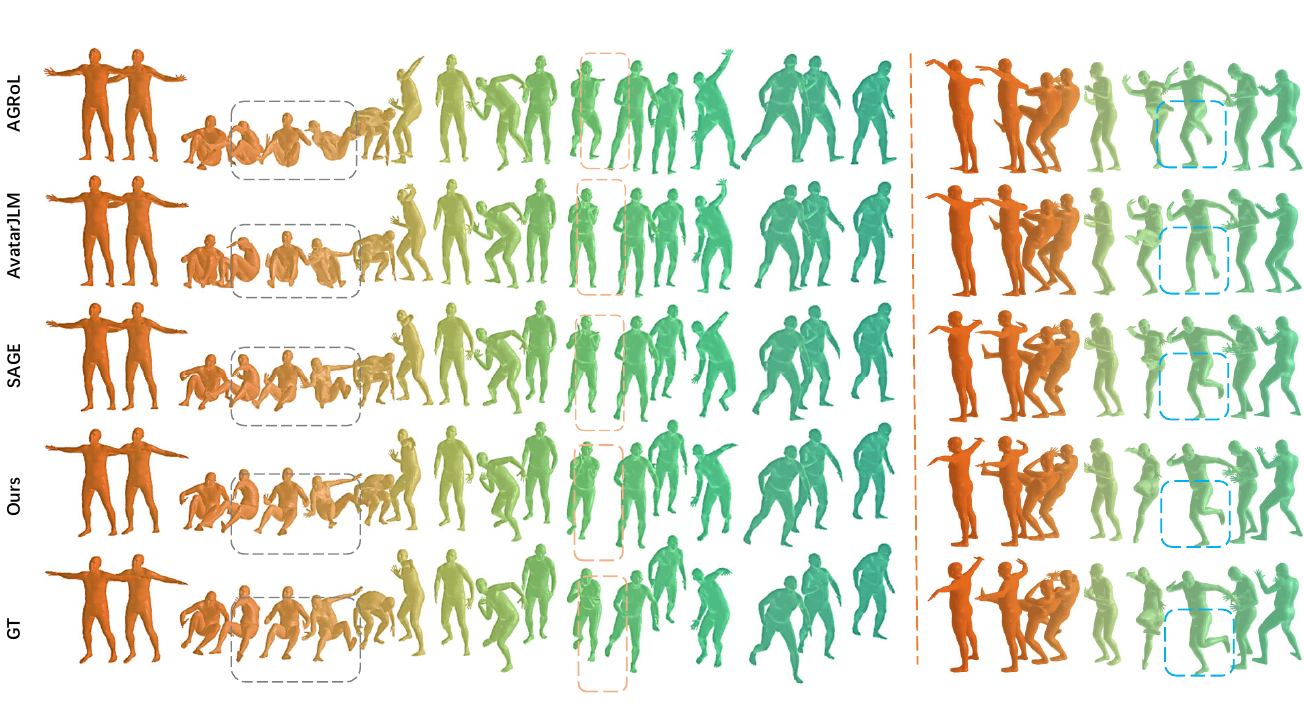}
    \caption{Visualization results of continuous pose sequences compared with other methods. Dashed boxes are used to highlight the most significant differences.}
    \label{fig3}
\end{figure*}

\subsection{Evaluation Metrics} 
Consistent with previous studies~\cite{3,6,7,38,12}, we evaluate performance using three categories of metrics. The first category measures the relative rotation error and includes the Mean Per Joint Rotation Error (MPJRE) [degrees]. The second category focuses on velocity-related metrics, reflecting motion smoothness, and includes the Mean Per Joint Velocity Error (MPJVE) [cm/s] and Jitter ($10^{2}$$m/s^{3}$)~\cite{12}. The third category evaluates position-related errors and includes the Mean Per Joint Position Error (MPJPE) [cm], Root PE, Hand PE, Upper PE, and Lower PE.

\subsection{Evaluation Results} 
To compare with the latest works~\cite{3,50,6,2,5,7}, the proposed model is evaluated on the AMASS~\cite{10} dataset with two different settings. In the first setting, we follow the approach of ~\cite{39,3,5,6}, which uses three subsets of the AMASS dataset, including CMU~\cite{33}, BMLrub~\cite{34}, and HDM05~\cite{35}. In the second setting, following~\cite{7}, the performance of the proposed model is further validated on a larger dataset from AMASS, including 14 subsets~\cite{33,34,35,40,41,42,43,44,45,47,48,49,47,29}. The datasets in both settings are randomly divided into training and test sets with 90\% and 10\% of the data, respectively.


\begin{table}[h]
\centering
\caption{Ablation study on the effects of different components in the SSD-Poser.}
\label{tab:Table4}
\begin{tabular}{l|cccc}
\toprule
Method & MPJRE & MPJPE & MPJVE & Jitter \\
\midrule
w/o PSSB & 2.49 & 3.27 & 20.55 & 9.50 \\
w/o SM block & 2.80 & 3.90 & 29.04 & 15.50 \\
w/o L2 norm loss & 2.66 & 3.61 & 25.53 & 12.09 \\
w/o FAD & 2.43 & 3.17 & 27.80 & 23.22 \\
\midrule\rowcolor[gray]{0.94} 
\textbf{SSD-Poser (Ours)} & \textbf{2.41} & \textbf{3.15} & \textbf{19.32} & \textbf{8.19} \\
\bottomrule
\end{tabular}
\end{table}

\begin{figure*}[h]
    \centering
    \includegraphics[width=0.9\linewidth]{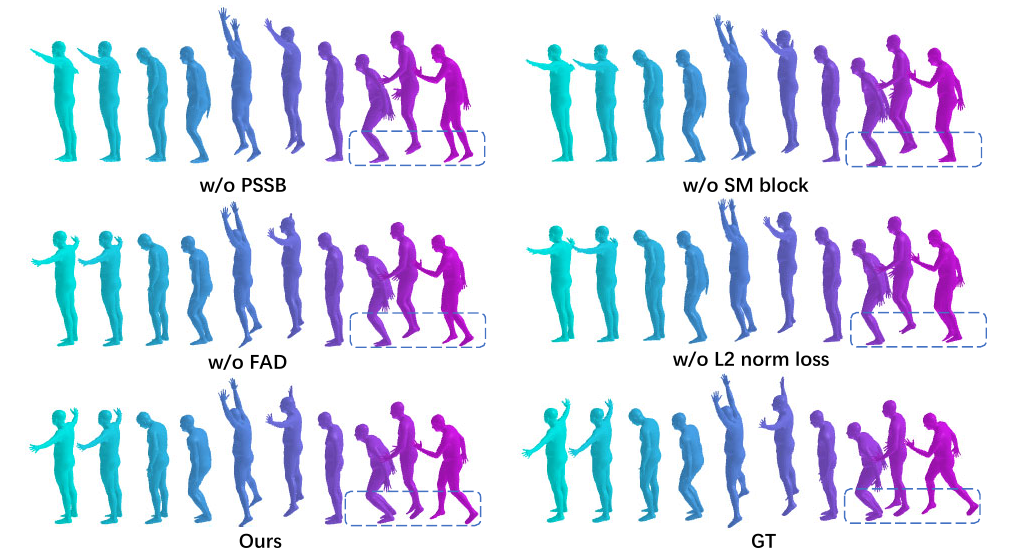}
    \caption{Visualization results of continuous pose sequences compared with ablation methods. Dashed boxes are used to highlight the most significant differences.}
    \label{fig4}
\end{figure*}

In the first setting, Table \ref{tab:Table1} represents that the proposed method achieves the best overall performance across methods. Specifically, the rotation-related and position-related metrics have a significant reduction compared to others, demonstrating the superiority of the proposed method for full-body pose estimation. For velocity-related metrics, while AGRoL demonstrates competitive results in MPJVE and SAGE achieves slightly lower jitter, their stacked generative blocks result in extra computational demands and slower inference speeds (shown in Table~\ref{tab:table2}). 

Moreover, in the second setting, as shown in Table \ref{tab:Table3}, the proposed model remarkably outperforms other methods. Notably, the proposed model achieves superior results across three key metrics, the 9.54\% reduction in MPJRE, the 9.49\% decrease in MPJPE, and the 9.97\% improvement in MPJVE compared to SAGE~\cite{7}. 

\begin{table*}[h]
\centering
\caption{Ablation study on the effects of different input sequence lengths in SSD-Poser.}
\label{tab:Table5}
\begin{tabularx}{\textwidth}{l|>{\centering\arraybackslash}X>{\centering\arraybackslash}X>{\centering\arraybackslash}X>{\centering\arraybackslash}X>{\centering\arraybackslash}X>{\centering\arraybackslash}X>{\centering\arraybackslash}X>{\centering\arraybackslash}X}
\toprule
Input Sequence Length & MPJRE & MPJPE & MPJVE & Hand PE & Upper PE & Lower PE & Root PE & Jitter \\
\midrule
41 & 2.41 & 3.21 & 23.33 & 1.35 & 1.40 & 5.83 & 2.88 & 13.04 \\
\rowcolor[gray]{0.94} 
\textbf{96 (Ours)} & \textbf{2.41} & \textbf{3.15} & 19.32 & \textbf{1.17} & \textbf{1.34} & \textbf{5.78} & \textbf{2.82} & 8.19 \\
144 & 2.49 & 3.31 & 19.54 & 1.28 & 1.41 & 6.04 & 2.93 & 7.47 \\
196 & 2.57 & 3.40 & \textbf{19.28} & 1.31 & 1.45 & 6.21 & 3.04 & 6.88 \\
225 & 2.64 & 3.50 & 19.61 & 1.27 & 1.48 & 6.41 & 3.11 & \textbf{6.65} \\
\bottomrule
\end{tabularx}
\end{table*}

\begin{table}[h]
\centering
\caption{Ablation study on the effects of different number of blocks in SSAEs.}
\label{tab:Table6}
\begin{tabular}{l|ccccccccc}
\toprule SSAE Blocks & Params(M) & MPJRE & MPJPE & MPJVE  & Jitter \\
\midrule
3 & 5.62 & 2.43 & 3.20 & 20.19  & 8.87 \\
\rowcolor[gray]{0.94} 
\textbf{4 (Ours)} & 7.34 & 2.41 & 3.15 & \textbf{19.32} & \textbf{8.19} \\ 
5 & 9.07 & \textbf{2.37} & \textbf{3.11} & 19.54 & 8.63 \\
\bottomrule
\end{tabular}
\end{table}



The proposed SSD-Poser not only delivers high accuracy but also ensures exceptional inference speed. For a fair comparison, all inference times are measured on the same NVIDIA 4090 device and the same data,  which includes 536 distinct motion sequences from the test dataset.  As shown in Table \ref{tab:table2}, the average inference time per sequence and the number of model parameters are calculated for each approach. The proposed model processes a single motion sequence in just $0.007s$, showing outstanding speed performance. This impressive real-time capability underscores the model’s suitability for AR/VR scenarios, effectively addressing user demands for responsiveness.

Visualization results are presented in Fig \ref{fig2} and Fig \ref{fig3}. The Fig \ref{fig2} presents 3D human reconstruction results for various actions, comparing our proposed method with other methods. Red regions indicate reconstruction errors, where wider or deeper red areas represent larger difference. These visuals validate the robustness of the proposed model, especially in predicting lower body movements, such as the squatting pose (in the second line), the side-kick pose (in the third line), .

Continuous motion sequence results are shown in Fig \ref{fig3}, with samples taken every 120 frames. Overall, the proposed method achieves superior reconstruction fidelity and smoothness. For instance, as highlighted by orange dashed boxes, the proposed method accurately lifts the right leg, closely matching the ground truth (GT). Additionally, as marked by the gray dashed boxes, other methods exhibit severe deformations during complex grounding actions, whereas the proposed method remains stable, particularly for the middle avatar.


\subsection{Ablation Study} 
In this section, the ablation studies are conducted on the AMASS dataset~\cite{10} following the first setting to verify the impact of each component and parameter in our designed model. The quantitative results are shown in Table~\ref{tab:Table4}-Table~\ref{tab:Table6}.

{\textbf{Effects of different components in SSD-Poser.}} To verify the indispensable role of each component and the loss function, multiple ablation experiments are conducted, as shown in Table \ref{tab:Table4} and Fig \ref{fig4}. These results demonstrate that the performance deteriorates when any component is removed. Fig \ref{fig4} visualizes the significant contributions of each component, particularly in actions like running and jumping, as indicated by the blue dashed box. Moreover, in Table \ref{tab:Table4}, it is noteworthy that when deliver ablation study on the FAD module, most metrics remain relatively stable. However, the velocity-related metrics of MPJVE and Jitter deteriorate significantly, indicating a loss of high-frequency motion capture. 

{\textbf{Effects of different input sequence lengths in SSD-Poser.}} The proposed SSD-Poser model takes sparse tracking signals with different sequence lengths as input, and the results are shown in Table~\ref{tab:Table5}. In general, SSD-Poser achieves the best overall performance when the input length sequence T=96. Additionally, Table~\ref{tab:Table5} demonstrates that when the sequence length exceeds 96, jitter begins to decrease with longer sequence, but both MPJRE and MPJPE increase significantly, indicating a detrimental effect on the accuracy of human pose estimation. Furthermore, the results show that when the sequence length is shorter than 96, such as T=41, all metrics deteriorate noticeably, except for MPJRE.

{\textbf{Effects of different number of blocks in SSAEs.}} Aiming to determine the optimal number of SSAE blocks, we evaluate the model’s performance with varying numbers of SSAE blocks, as shown in Table~\ref{tab:Table6}. Considering precision, smoothness, and model efficiency, the SSAEs network is established with four SSAE blocks.



\section{Conclusion}
In this paper, a lightweight yet effective model, SSD-Poser, is proposed for full-body pose estimation from sparse observations. In the model, SSAEs is designed to efficiently capture spatiotemporal dependencies while reducing computational overhead. The introduction of FAD handles variable-frequency signals and refines body poses, achieving higher realism. Extensive experiments demonstrate that the proposed model surpasses state-of-the-art approaches in reconstruction accuracy while achieving superior inference speed within a lightweight architecture. The proposed method has significant potential to enhance user experiences in AR/VR applications by providing seamless and responsive motion tracking.

\newpage

\bibliographystyle{ACM-Reference-Format}
\bibliography{references.bib}

\appendix

\end{document}